\documentclass[sigconf]{acmart}
\usepackage{subfigure}
\usepackage{CJKutf8}
\usepackage{extarrows}
\usepackage{multirow}
\usepackage{hyperref}
\usepackage{url}
\usepackage{booktabs}

\AtBeginDocument{%
  \providecommand\BibTeX{{%
    \normalfont B\kern-0.5em{\scshape i\kern-0.25em b}\kern-0.8em\TeX}}}

\copyrightyear{2023}
\acmYear{2023}
\setcopyright{acmlicensed}\acmConference[WWW '23]{Proceedings of the ACM Web Conference 2023}{May 1--5, 2023}{Austin, TX, USA}
\acmBooktitle{Proceedings of the ACM Web Conference 2023 (WWW '23), May 1--5, 2023, Austin, TX, USA}
\acmPrice{15.00}
\acmDOI{10.1145/3543507.3583328}
\acmISBN{978-1-4503-9416-1/23/04}

\begin{document}

\title{Attribute-Consistent Knowledge Graph Representation Learning for Multi-Modal Entity Alignment}


\author{Qian Li}
\affiliation{
  \institution{School of Computer Science and Engineering, BDBC,\\ Beihang University}
  \state{Beijing}
  \country{China}}
\email{liqian@act.buaa.edu.cn}

\author{Shu Guo}
\affiliation{
  \institution{National Computer Network Emergency Response Technical Team/Coordination Center of China}
  \state{Beijing}
  \country{China}}
\email{guoshu@cert.org.cn}

\author{Yangyifei Luo}
\affiliation{
  \institution{School of Computer Science and Engineering, BDBC,\\ Beihang University}
  \state{Beijing}
  \country{China}}
\email{luoyangyifei@buaa.edu.cn}

\author{Cheng Ji}
\affiliation{
  \institution{School of Computer Science and Engineering, BDBC,\\ Beihang University}
  \state{Beijing}
  \country{China}}
\email{jicheng@act.buaa.edu.cn}

\author{Lihong Wang}
\affiliation{
  \institution{National Computer Network Emergency Response Technical Team/Coordination Center of China}
  \state{Beijing}
  \country{China}}
\email{wlh@cert.org.cn}

\author{Jiawei Sheng}
\affiliation{
  \institution{Institute of Information Engineering, Chinese Academy of Sciences, \\School of Cyber Security, UCAS }
  \state{Beijing}
  \country{China}}
\email{shengjiawei@iie.ac.cn}

\author{Jianxin Li}
\authornote{Corresponding author}
\affiliation{
  \institution{SCSE, Beihang University, \\Zhongguancun Lab}
  \state{Beijing}
  \country{China}}
\email{lijx@act.buaa.edu.cn}






\renewcommand{\shortauthors}{Qian Li, et al.}

\begin{abstract}
The multi-modal entity alignment (MMEA) aims to find all equivalent entity pairs between multi-modal knowledge graphs (MMKGs). Rich attributes and neighboring entities are valuable for the alignment task, but existing works ignore contextual gap problems that the aligned entities have different numbers of attributes on specific modality when learning entity representations.
In this paper, we propose a novel attribute-consistent knowledge graph representation learning framework for MMEA (ACK-MMEA) to compensate the contextual gaps through incorporating consistent alignment knowledge. Attribute-consistent KGs (ACKGs) are first constructed via multi-modal attribute uniformization with merge and generate operators so that each entity has one and only one uniform feature in each modality. The ACKGs are then fed into a relation-aware graph neural network with random dropouts, to obtain aggregated relation representations and robust entity representations. In order to evaluate the ACK-MMEA facilitated for entity alignment, we specially design a joint alignment loss for both entity and attribute evaluation. Extensive experiments conducted on two benchmark datasets show that our approach achieves excellent performance compared to its competitors.
\end{abstract}

\ccsdesc[500]{Computing methodologies~Knowledge representation and reasoning}
\ccsdesc[500]{Computing methodologies~Natural language processing}
\ccsdesc[500]{Theory of computation~Semantics and reasoning}
\keywords{Entity alignment, Multi-modal knowledge graph representation}



\maketitle

\section{Introduction}\label{sec:intro}

Knowledge graphs (KGs) have become a popular data structure for representing factual knowledge in form of RDF triples. Recently, there is a growing trend to incorporate multi-modal information into KGs, i.e., Multi-Modal Knowledge Graphs (MMKGs), which support various cross-modal tasks, e.g., recommendation systems~\cite{DBLP:conf/cikm/SunCZWZZWZ20, DBLP:conf/cikm/XuCLSSZZZZ21} and question answering systems ~\cite{DBLP:journals/corr/abs-2204-09220, DBLP:journals/corr/abs-2212-05767}. However, MMKGs often suffer from low coverage and incompleteness. To improve the coverage of these MMKGs, a viable approach termed as multi-modal entity alignment (MMEA) is proposed to identify the equivalent entity pairs (i.e., alignment seeds) in different MMKGs, by integrating the attribute information of text and image. In this way, MMKGs can obtain useful knowledge from other KG. 

Although the rich attributes and neighboring entities in MMKGs provide valuable pieces of evidence for MMEA~\cite{DBLP:conf/esws/LiuLGNOR19}, the inevitable heterogeneity of MMKGs makes it difficult to learn and fuse knowledge representations from distinct modalities. A series of effective methods have been developed to conquer these challenges, and the detailed description is in Appendix~\ref{sec:appe-Related}. The PoE method~\cite{DBLP:conf/esws/LiuLGNOR19} composited representations of entities by concatenating all modality features, which could not capture the potential interactions among heterogeneous modalities and therefore limited the performance of MMEA. Later works~\cite{DBLP:conf/ksem/ChenLWXWC20,DBLP:journals/ijon/GuoTZZL21} designed multi-modal fusion modules to properly integrate attributes and entities, in order to better predict alignments according to aggregated embeddings. All of these methods would learn entity representations by harnessing their whole associated attributes and neighboring entities. Nevertheless, they ignore the contextual gaps between entity pairs and in turn constrain the effectiveness of the entity alignment.


\begin{figure}[t]
\centering
\includegraphics[width=\linewidth]{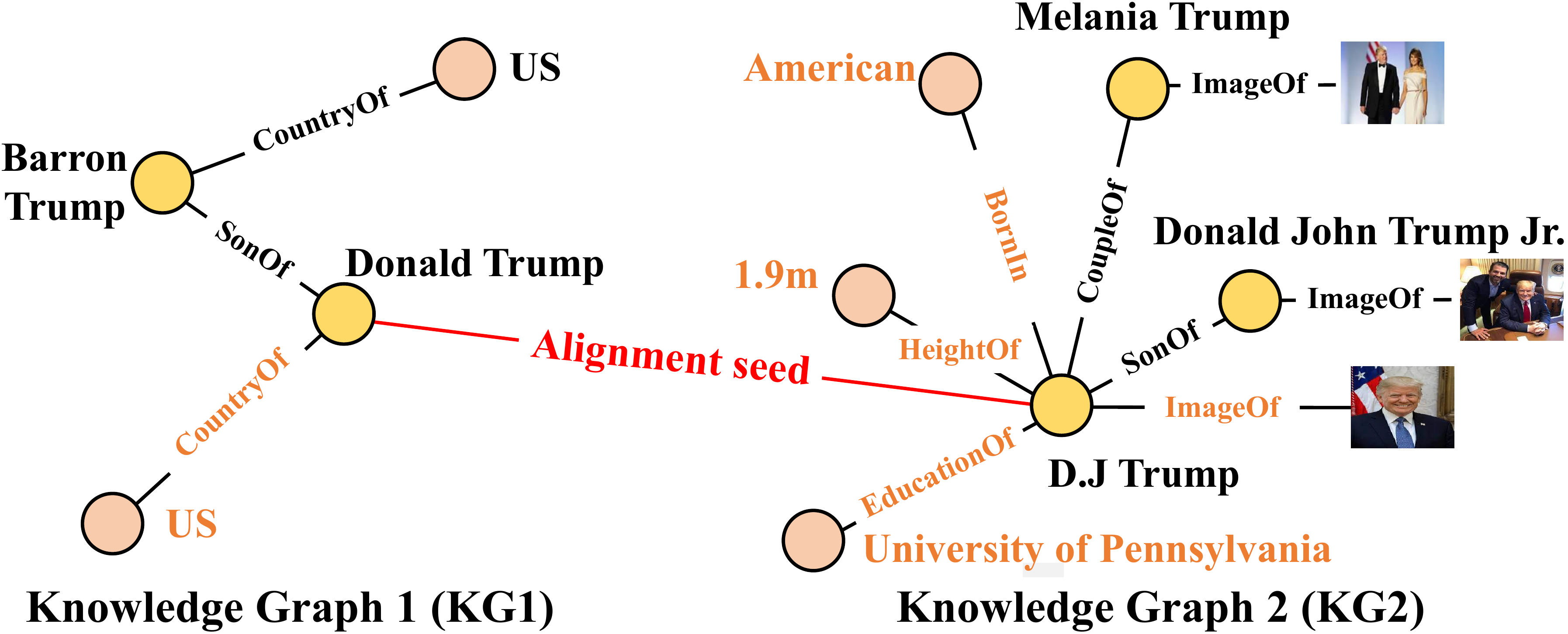}
\caption{An example of the MMEA task between KG1 and KG2. The yellow and orange circles are the entity node and attribute nodes, respectively.}
\label{fig:define}
\end{figure}

The contextual gap, which means entities may associate with different numbers of attributes or even lack some modalities, is inevitable due to information redundancy or absence. Such inconsistencies between the equivalent entity pairs make alignments error-prone. Figure~\ref{fig:define} illustrates a toy example of the contextual gap in MMEA. (a) Difference in the number of attributes. The entity \textit{Donald Trump} in KG1 associates with only one text attribute \textit{US}. With such limited contextual information, it is not easy to determine the identity of entity \textit{Donald Trump}. In contrast, the identity of entity \textit{D.J Trump} in KG2 will be more specific as he contains richer text attributes, which 
causes that similarity of text attributes (between \textit{US} and \textit{American}) is diluted by existing aggregation-based approaches. 
The contextual gaps caused by different numbers of text attributes make it hard to obtain the attribute-consistent alignment knowledge and judge that both actually refer to the same real-world identity.
(b) Lack of modal attribute. Missing attributes also leads to contextual gap problems since unique attributes are neglected for alignment on existing fusion baselines, as such missing image attribute of \textit{Donald Trump} makes \textit{D.J Trump} difficult to align with it.
To overcome the above challenges, we propose a novel \underline{\textbf{M}}ulti-\underline{\textbf{M}}odal \underline{\textbf{E}}ntity \underline{\textbf{A}}lignment framework based on \underline{\textbf{A}}ttribute-\underline{\textbf{C}}onsistent \underline{\textbf{K}}nowledge graph representation learning
, termed as ACK-MMEA\footnote{The source code is available at \url{https://github.com/xiaoqian19940510/ACK-MMEA}.}. Specifically, for the attribute information, we design a multi-modal attribute uniformization method to obtain attribute-consistent KGs (ACKGs) with one uniform attribute for each modality of all entities. That means in the ACKGs every entity will possess only one attribute for each modality (i.e., one text and one image attribute). To generate such an attribute-consistent MMKG, we devise merge and generate operators for each entity to deal with the attribute redundancy and absence respectively. The former is to compress associated multiple attributes into one with an attention-based approach to filter out the noise attributes, while the latter is to expand a new attribute if the entity has no attribute in specific modality. We further devise a ConsistGNN to enforce consistent attribute aggregation for entity representation. Given a relation triple in the above ACKGs, ConsistGNN first obtains an aggregated relation representation by simultaneously integrating relational features under every modality (i.e., entity/text/image). Then entity representations are obtained using a relation-aware entity encoder. As the newly constructed ACKGs may introduce noises, random dropouts on neighbors are employed to produce more robust representations. Finally, we design a joint alignment loss for entity and attribute evaluation. Experimental results show that our approach can achieve excellent performance on two MMEA benchmark datasets.
Our contributions are summarized as follows: 
\begin{itemize}
    \item We propose a novel multi-modal entity alignment framework to incorporate the consistent alignment knowledge through leveraging an attribute-consistent knowledge graph representation learning method. To the best of our knowledge, this is the first work to tackle the contextual gap problems in the MMEA task.
    \item We design a multi-modal attribute uniformization method using merge and generate operators to derive an attribute-consistent MMKG, and a ConsistGNN model to aggregate consistent information.
    \item Experimental results indicate the framework achieves state-of-the-art performance on two public MMEA datasets.
\end{itemize}

\begin{figure*}[t]
\includegraphics[width=\linewidth]{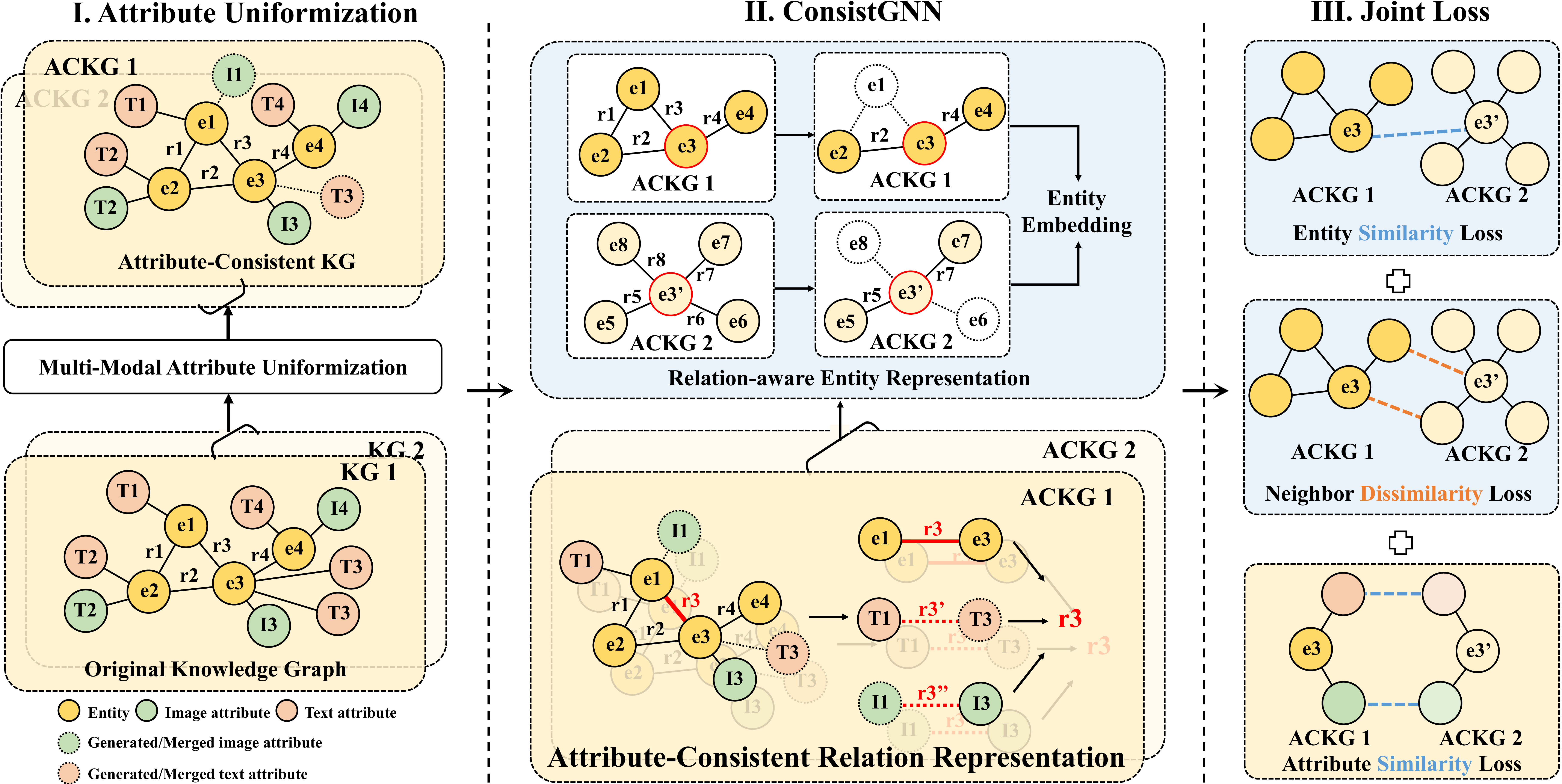}
\caption{The framework of ACK-MMEA. (I) Two new attribute-consistent MMKGs ($ACKG_1$ and $ACKG_2$) are generated by performing the multi-modal attribute uniformization on the original ones ($KG_1$ and $KG_2$). (II) ConsistGNN: Relation-aware GNN with dropouts is to aggregate consistent attributes and learn robust representations of entities. (III) Joint alignment loss with three objectives is used for parameter optimization.}
\label{fig:Framework}
\end{figure*}

\section{Preliminaries}
We first provide the definitions of multi-modal knowledge graph (MMKG) and multi-modal entity alignment (MMEA) as follows.

\paragraph{\textbf{Definition of MMKG}} 
A multi-modal knowledge graph, denoted as $KG =(\mathcal{E},  \mathcal{R}, \mathcal{A})$, is composed of relations between entities and associated attributes. Specifically, $\mathcal{E}, \mathcal{R}, \mathcal{A}$ are the sets of entities, relations, and multi-modal attributes, respectively, with size of $n_E, n_R, n_A$. We suppose that a $KG$ has two kinds of attributes, i.e., the text attributes $\mathcal{T}$ and image attributes $\mathcal{I}$.

In this paper, we aims to resolve the problem of contextual gaps for the entity alignment task on MMKG. In general, the entity alignment task includes cross-language entity alignment and multi-source entity alignment~\cite{DBLP:journals/aiopen/ZengLHLF21}. Following previous MMEA studies~\cite{DBLP:conf/ksem/ChenLWXWC20, DBLP:journals/ijon/GuoTZZL21, DBLP:conf/aaai/0001CRC21}, this paper focuses on the latter one.

\paragraph{\textbf{Definition of MMEA Task}} 
The multi-modal entity alignment task~\cite{DBLP:conf/ksem/ChenLWXWC20, DBLP:journals/ijon/GuoTZZL21, DBLP:conf/aaai/0001CRC21} is to identify whether a pair of entities in two multi-modal knowledge graphs is equivalent or not. 
Concretely, given two multi-modal knowledge graphs $KG_1$ and $KG_2$ with a pair of entity alignment seed $(v,{v}^{\prime})$, where $v$ and ${v}^{\prime}$ are entities in $KG_1$ and $KG_2$, the multi-modal entity alignment aims to identify whether they are equivalent. The main procedure is to learn entity representations in two multi-modal knowledge graphs and calculate the similarity between alignment seed $(v,{v}^{\prime})$. The set of entity alignment seeds is $\mathcal{S}=\left\{\left(v, v^{\prime}\right) \mid v \in \mathcal{E}, v^{\prime} \in \mathcal{E}^{\prime}, v \equiv v^{\prime}\right\}$.


\section{Framework}

This section introduces our proposed framework ACK-MMEA. 
As shown in Figure~\ref{fig:Framework}, ACK-MMEA consists of the following three modules: attribute uniformization, ConsistGNN and joint alignment loss. 
Firstly, the attribute uniformization module generates consistent attributes for each entity, wherein each entity has one attribute for every modality, respectively. 
Then, the ConsistGNN maps entities, attributes, and relations to a common representation space and learns heterogeneous relation as well as robust entity representations by aggregating consistent multi-modal attributes. 
Finally, the joint alignment loss combines the losses of entity similarity, attribute similarity, and neighbor dissimilarity to comprehensively evaluate the attribute-consistent MMKG. 

\subsection{Multi-Modal Attribute Uniformization}\label{Section 4.1}

To better tackle the inconsistency issue of attributes in MMKG, we divide the original multi-modal knowledge graph $KG$ into multiple knowledge graphs under each modality. 

An MMKG contains text and image attributes, which can be divided into three knowledge graphs $\{KG_E, KG_T, KG_I\}$, corresponding to the entity, text and image graphs, respectively. For each divided knowledge graph $KG_X$, where $X \in \{E,T,I\}$, the representations of nodes $\mathbf{E}_{X}$ (entities, texts, images) are :
\begin{align}
\footnotesize
    \mathbf{E}_{\!X}=\mathbf{F}_{X} \cdot \mathbf{W}_{\!X}, 
\end{align}
where $\mathbf{F}_{\!X}$ are the initial representations of nodes. For each modality of nodes, (a) in $KG_E$, entity $\mathbf{F}_{E} \in \mathbb{R}^{n_E \times d_E}$ is initialized by the TransE model~\cite{DBLP:conf/nips/BordesUGWY13}, with $d_E$ as its dimension; (b) in $KG_T$, text attribute $\mathbf{F}_{T} \in \mathbb{R}^{n_T \times d_T}$ is initialized by the BERT~\cite{DBLP:conf/nips/VaswaniSPUJGKP17}, with $d_T$ as its dimension; (c) in $KG_I$, image attribute $\mathbf{F}_{I} \in \mathbb{R}^{n_I \times d_I}$ is initialized by the VGG16~\cite{DBLP:journals/corr/SimonyanZ14a}, with $d_I$ as its dimension. $\mathbf{W}_{\!E} \in \mathbb{R}^{d_E \times d}$, $\mathbf{W}_{\!T} \in \mathbb{R}^{d_T \times d}$, $\mathbf{W}_{\!I} \in \mathbb{R}^{d_I \times d}$ are learnable transformation matrix, mapping the initial representations of different types of nodes into a common $d$-dimensional space. 

\begin{figure}[t]
\includegraphics[width=\linewidth]{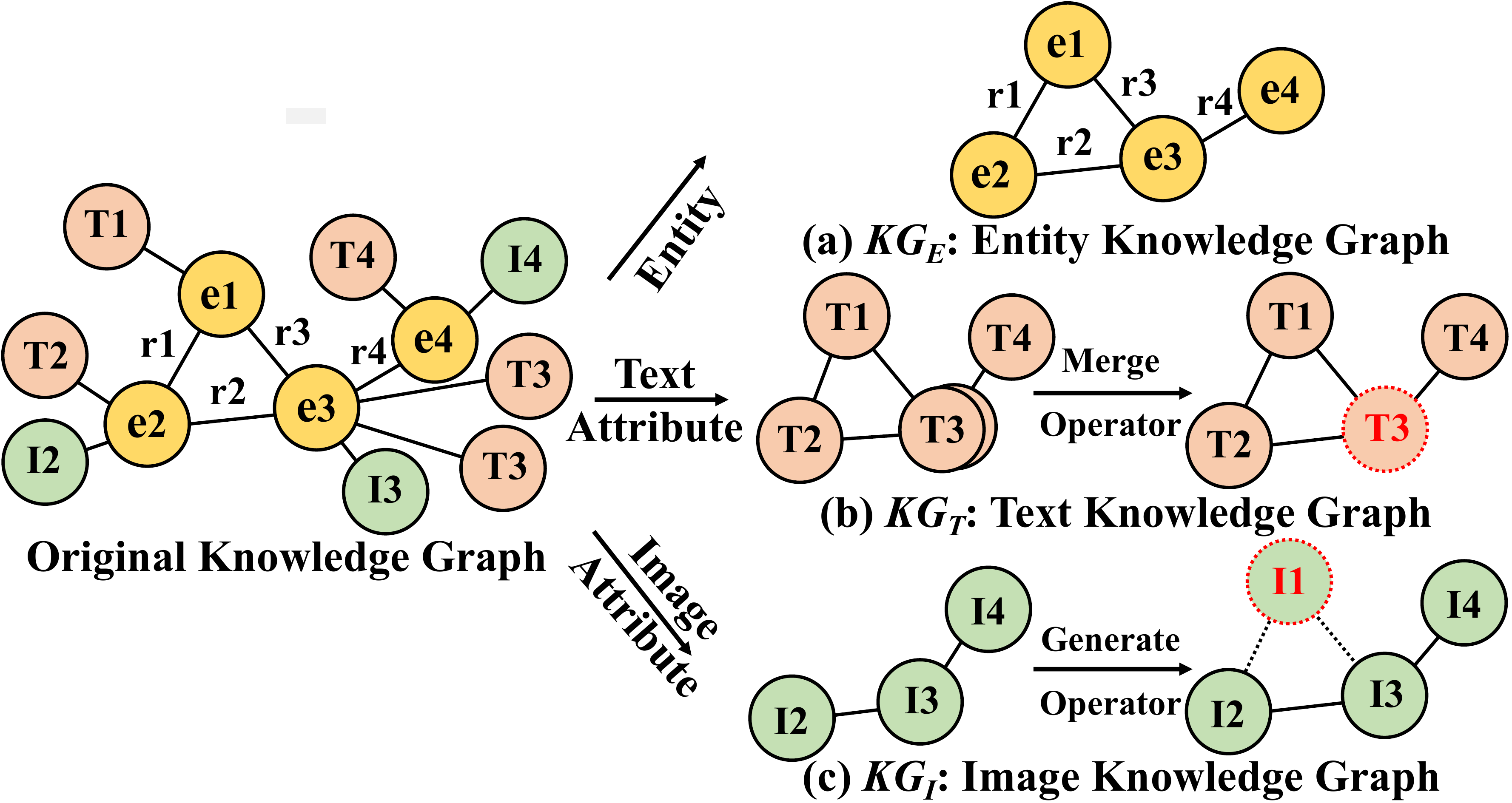}
\caption{Schematic diagram of the multi-modal attribute uniformization. (a) is the entity knowledge graph only including entity. (b) and (c) are the text and image knowledge graphs, connected relying on the relation of (a). We use the merge and generate operators to make the text and image attributes uniformization.}
\label{fig:generator}
\end{figure}

It is noticed that attributes of entities in the original MMKG are inconsistent, as shown in Figure~\ref{fig:generator}. 
The inconsistency means that the entities in each pair have different number of attributes for a specific modality as discussed in Section~\ref{sec:intro}. Such contextual gaps require that the model has the ability to select and generate attributes that contribute to the MMEA task. However, it is hard to determine which and how many attributes to use for the entity alignment, since the contextual gap has no particular structural pattern and different entities have different severity of problems.

To this end, we want to seek generic solutions to contextual gap problem in this paper. We implement the following two uniformization operators on the original MMKG to map entities, attributes, and their relations to a common representation space as well as ensure consistency among them simultaneously. (a) Merge Operator. We aggregate all attributes of each modality into one through an attention-based mechanism, to represent the uniform feature of the entity in the specific modality that is helpful to the MMEA task. (b) Generate Operator. We propose to use the neighbors' attributes to generate the missing attribute. The combination of merge and generate operators alleviates the problem of contextual gap.

\paragraph{\textbf{Merge Operator.}} For MMKGs, there may be multiple attributes $\{\mathbf{e}_{v,A,i}\}_{i=1}^{n_v}$ of entity $v \in \mathcal{E}$ in $\mathbf{E}_{A}$, where $A \in \{T, I\}$, and $n_{v}$ is the number of attribute in $\mathbf{h}_{v,A}$. In order to aggregate information across the attributes in the same modality, we perform a merge operator for each modality.
The operator is implemented by a learnable graph attention~\cite{DBLP:conf/emnlp/LiuCPLC20} which can discard redundant contextual information with lower weights, making the final consistent attribute helpful to the MMEA task.
Thus we can obtain an aggregated attribute $\mathbf{E}_{A}^0$ which compresses redundant contextual information:
\begin{equation}
\mathbf{E}_{A}^{(0)}[v]=\sigma(\sum_{i=1}^{n_v} \alpha_{i} \cdot \mathbf{W}_{\!M}  \mathbf{e}_{v,A,i}),
\end{equation}
where $\alpha_{i}$ is the learned attention weight for $i$-th attribute, $\mathbf{W}_{\!M} \in \mathbb{R}^{d \times d}$ is a learnable parameter, $\sigma$ is $\text{ReLU}(\cdot)$ function.

\paragraph{\textbf{Generate Operator.}}
For the MMKG, many entities miss their attributes of the specific modality. It is observed that the neighbors' attributes in the same modality usually provide helpful information to generate the attribute of the target entity. Intuitively, the image attribute of the entity "Donald Trump" is similar to their children's, which means their representations are close. Therefore, to compensate the contextual gap caused by missing attributes, we generate the attribute $\mathbf{E}_{A}^{(0)}$ with an average aggregation of only the first-order neighbor attributes. 
\begin{align}
\footnotesize
    \mathbf{E}_{A}^{(0)}[v] = \sigma\left(\mathbf{W}_{G} \cdot \mathrm{MEAN}\left(\{\mathbf{e}_{u,A} | u \in \mathcal{N}\!\left({v}\right)\}\right)\right.,
\end{align}
where $\mathbf{W}_{G} \in \mathbb{R}^{d \times d}$ is a learnable transformation matrix, $\mathcal{N}(v)$ is the first-order neighbor set of the entity $v$, $\sigma$ is the $\text{ReLU}(\cdot)$ function, and $\mathbf{e}_{u,A}=\mathbf{E}_{A}[u]$. 


In this way, the contextual gaps can be relieved, as the attributes of entities in KGs are consistent via the two uniformization operators. Such attribute-consistent KGs (ACKGs) would be helpful to balanced attribute integration during entity representation learning, leading to more accurate alignments.

\subsection{ConsistGNN}
We further design a GNN model named ConsistGNN to derive relation and entity representations respectively based on attribute-consistent relation representation encoder and relation-aware entity representation encoder, enabling consistent modality aggregation on attribute-consistent KGs.

In ACKGs, there is only one attribute for every entity under each modality. It makes sense that the relations between the same modality of attributes can be somewhat analogous to those between entities. Thus, we define the representations of entity relations $\mathbf{R}^{(0)}$ and attribute relations $\mathbf{R}_{A}^{(0)}$ are calculated as follows respectively:
\begin{align}
\footnotesize
    \mathbf{R}^{(0)}=\mathbf{R} \cdot \mathbf{W_0}, \mathbf{R}_{A}^{(0)}=\mathbf{R}_{A} \cdot \mathbf{W}_{0,A}, 
\end{align}
where $\mathbf{R}, \mathbf{R}_{A} \in \mathbb{R}^{n_r \times d_E}$ are the initial representations of relations, which are calculated from TransE. Specifically, for two connected entities $(h,t)$, the attribute relation $\mathbf{r}_{ht,A}=|\mathbf{e}_{t,A}-\mathbf{e}_{h,A}| \in \mathbf{R}_{A}$ is calculated by tail attribute $\mathbf{e}_{t,A}$ and head attribute $\mathbf{e}_{h,A}$. $\mathbf{W}_0, \mathbf{W}_{0,A} \in \mathbb{R}^{d_E \times d}$ are learnable transformation matrix for mapping initial relation representations into the common space.

To obtain entity representations containing the consistent attribute information, we first initialize the entity representations of the ACKGs as follows:
\begin{equation}
\mathbf{E}_{E}^{(0)}=\sigma(\mathbf{E}_{E}\mathbf{W}_{\!c,E} + \sum_{A \in \{T,I\}} \mathbf{E}_{A}^{(0)} \mathbf{W}_{\!c,A}),
\end{equation}
where $\mathbf{W}_{\!c,E},\mathbf{W}_{\!c,A}\in \mathbb{R}^{d \times d}$ are learnable transformation matrices. We then feed the initial entity and relation representations into ConsistGNN equipped with attribute-consistent relation representation encoder and relation-aware entity representation encoder. Specifically, we calculate the representations of nodes $\mathbf{E}^{(l)}=\{\mathbf{E}_{E}^{(l)}, \mathbf{E}_{A}^{(l)}\}$ and relations $\mathbf{R}^{(l)}$ in the $l$-th layer as follows:
\begin{align}
\footnotesize
    \mathbf{E}^{(l)}, \mathbf{R}^{(l)}\!\!=\!\operatorname{ConsistGNN}\left(\mathbf{E}^{(l-1)},\! \mathbf{R}^{(l-1)}\right).
\end{align}

\paragraph{\textbf{Attribute-Consistent Relation Representation.}}

The attribute uniformization obtains unique identity of each modality for every entity. The relation of two entities is close to the attribute relations of theirs. Thus, we propose to use the consistent attribute information and utilize the attribute relation for entity relation learning.
Specifically, we propose an attribute-consistent relation representation encoder for utilizing the multi-modal attribute information, where the relation of two attributes is represented by the combination of themselves: 
\begin{align}
\footnotesize
    \mathbf{R}^{(l)}[u,v] & =\operatorname{ReLU}(\mathbf{W}_{E\!-\!E}^{(l)} \; \mathbf{r}_{uv}^{(l-1)}  \nonumber \\
    &+\! \sum_{A \in \{T,I\}}\! \mathbf{W}_{E\!-\!A}^{(l)} [\mathbf{e}^{(l-1)}_{u,A} || \mathbf{e}^{(l-1)}_{v,A}] ),
\end{align}
where $\mathbf{r}_{uv}^{(l-1)}=\mathbf{R}^{(l-1)}[u,v]$, and $\operatorname{ReLU(\cdot)}$ is the activation function.

\paragraph{\textbf{Relation-aware Entity Representation}}
To make the entity representation have fault tolerance ability for the generated ACKGs, we adopt random dropouts~\cite{DBLP:conf/nips/YouCSCWS20} on neighbors to improve the robustness of entity representation, which assumes that missing part of entities does not affect the semantic meaning of ACKG. For each entity $v$, we randomly discard certain portion of neighboring entities along with the relations connected to themselves. The neighbor set of each entity $v$ is therefore updated as: 
\begin{align}
\footnotesize
    \mathcal{N}(v) \leftarrow& \operatorname{drop}\left(\mathcal{N}(v); \rho \right) \nonumber \\
    =& \{u_i | u_i \in \mathcal{N}(v), p(i)=1, p(i) \sim \mathrm{Ber}(1-\rho)\},
\end{align}
where $\rho$ is the random dropouts rate. $\operatorname{drop}(\cdot)$ is the neighbors random dropouts function, and $\mathrm{Ber(\cdot)}$ is the Bernoulli distribution. The best random dropouts rate is 0.35.  
Afterwards, we update the entity representation with the neighbors random dropouts knowledge graph, utilizing the relation and attribute information for each entity.
The entity representations in the $l$-th layer are
\begin{align}
\footnotesize
    \mathbf{E}_{E}^{(l)}[v]&\!=\!\mathbf{W}_{h,E}[\mathbf{e}_{v,E}^{(l-1)} || \frac{1}{D_v} \!\! \sum_{u \in \mathcal{N}(v)} \!\! [\mathbf{e}_{u,E}^{(l-1)} || \mathbf{r}_{uv}^{(l)}]], 
\end{align}
where $D_v$ is the degree of entity $v$. 
The updates of attribute is the special case of entity updates. The attribute representations in the $l$-th layer $E_{A}^{l}[v]$ are the integration of attribute representations $E_{A}^{l-1}[v]$ and entity representations $E_{E}^{l-1}[v]$ in ($l$-$1$)-th layer.

\subsection{Joint Alignment Loss}
We design a joint alignment loss function, which utilizes the alignment losses of entity similarity, attribute similarity and neighbor dissimilarity to evaluate the attribute-consistent MMKG for multi-perspective assessment. 

\paragraph{\textbf{Aligned Entity Similarity}} The entity similarity constraint loss is as follows:
\begin{align}
\footnotesize
    \mathcal{L}_{i}^{EA}=\operatorname{sim}(\mathbf{e}_{i}, \mathbf{{e}}_{i}^{\prime})-\operatorname{sim}(\mathbf{e}_{i}, \mathbf{\overline{e}}_{i}^{\prime})-\operatorname{sim}(\mathbf{\overline{e}}_{i}, \mathbf{{e}}_{i}^{\prime}),
\end{align}
where $(\mathbf{e}_{i}, \mathbf{{e}}^{\prime}_{i})$ are the final representations of aligned seeds $(v_i,{v^{\prime}_i})$ of $KG_{1}$ and $KG_{2}$. $\mathbf{\overline{e}}_{i}$ and $\mathbf{\overline{e}}_{i}^{\prime}$ are the negative samples of the seeds. $\operatorname{sim(\cdot,\cdot)}$ is the cosine distance.

\paragraph{\textbf{Aligned Attribute Similarity}}
In addition, in order to make the relation between the same type of attributes similar to the two adjacent entities, we design the attribute similarity constraint loss as follows:
\begin{align}
\footnotesize
    \mathcal{L}^{attr}_i=\sum_{A \in \{T,I\}} \operatorname{sim}(\mathbf{e}_{i,A}, \mathbf{{e}}_{i,A}^{\prime}).
\end{align}

\paragraph{\textbf{Aligned Neighbor Dissimilarity}}
Furthermore, for more precise alignment of candidate entities, the entity ${v^{\prime}_i}$ in $KG_{2}$ should be characterized similar to entity $v_{i}$ in $KG_{1}$, and the neighbors $\mathcal{N}({v^{\prime}_i})$ of ${v^{\prime}_i}$ in $KG_{2}$  should be characterized dissimilar to entity $v_i$ in $KG_{1}$. Thus, we introduce the neighbor dissimilarity constraint loss, inspired by~\cite{DBLP:conf/sigir/WuWF0CLX21,DBLP:journals/corr/abs-2301-12104}:
\begin{align}
\footnotesize
    \mathcal{L}^{cont}_i\!\!=\!-\!\log\! \frac{\exp (\operatorname{sim}(\mathbf{e}_{i}, \! \mathbf{{e}}_{i}^{\prime}) / \tau\!)}{\sum_{{v^{\prime}\!_j} \in \mathcal{N}({v^{\prime}_i})} \!\exp (\operatorname{sim}(\mathbf{e}_{i},\! \mathbf{{e}}_j^{\prime})) / \tau\!)},
\end{align}
where $v^{\prime}_j$ is the neighbors of entity $v^{\prime}_i$, and $\tau$ is the temperature coefficient. 

The total alignment loss $\mathcal{L}$ is the weighted sum of three loss functions:
\begin{align}
\footnotesize
    \mathcal{L}=\lambda_{1} \mathcal{L}^{EA}+\lambda_{2} \mathcal{L}^{attr}+\lambda_{3} \mathcal{L}^{cont},
\end{align}
where $\lambda_{1}, \lambda_{2}, \lambda_{3}$ are the learnable hyper-parameters for joint alignment loss.


\begin{table*}[t]
\centering
\caption{Main experiments on FB15K-DB15K (top) and FB15K-YAGO15K (bottom) with different proportions of entity alignment seeds. The best results are highlighted in bold, and the underlined values are the second best result.  The "$\uparrow$" means the improvement compared to the second best result, and ``-" means that the results are not available.}
\resizebox{\linewidth}{!}{
\begin{tabular}{l|ccc|ccc|ccc}
\toprule
& \multicolumn{3}{c}{\textbf{FB15K-DB15K (20\%)}}  &  \multicolumn{3}{c}{\textbf{FB15K-DB15K (50\%)}}  &  \multicolumn{3}{c}{\textbf{FB15K-DB15K (80\%)}}   \\ 
\textbf{Methods}  & MRR (\%) & Hits@1 (\%) & Hits@10 (\%) & MRR (\%) & Hits@1 (\%) & Hits@10 (\%)   & MRR (\%) & Hits@1 (\%) & Hits@10 (\%)  \\ 
\midrule
\textbf{TransE} \cite{DBLP:conf/nips/BordesUGWY13} & 13.4  & 7.8  & 24.0     & 30.6 & 23.0 &  44.6  & 50.7   & 42.6 &  65.9 \\ 
\textbf{IPTransE}~\cite{DBLP:conf/ijcai/ZhuXLS17}    & 9.4   & 6.5   & 21.5  & 28.3 & 21.0 & 42.1    & 46.9  & 40.3 & 62.7 \\ 
\textbf{GCN-align} \cite{DBLP:conf/emnlp/WangLLZ18}   & 8.7  & 5.3  & 17.4    & 29.3  & 22.6  & 43.5  & 47.2  & 41.4  & 63.5  \\ 
 \textbf{SEA} \cite{DBLP:conf/www/PeiYHZ19}  & 25.5  & 17.0  & 42.5    & 47.0 & 37.3 &  65.7  & 50.5 &51.2 &78.4   \\  
 \textbf{IMUSE} \cite{DBLP:conf/dasfaa/HeLQ0LZ0ZC19} & 26.4  & 17.6  & 43.5    & 40.0 & 30.9 &  57.6  & 55.1  &45.7  &72.6 \\
  \textbf{AttrGNN}~\cite{DBLP:conf/emnlp/LiuCPLC20} & 34.3& 25.2 & 53.5   & \underline{54.7} & \underline{47.3} & \underline{72.1}  & 70.3 &  \underline{67.1} & 83.9  \\ \midrule
 \textbf{PoE} \cite{DBLP:conf/esws/LiuLGNOR19} & 17.0  & 12.6  & 25.1    & 53.3  &46.4  & 65.8  & \underline{72.1}  &66.6  & 82.0  \\ 
\textbf{PoE-rni} \cite{DBLP:conf/esws/LiuLGNOR19} & 28.3  & 23.2  & 39.0    & 44.2  & 38.0  & 55.7   & 55.8  & 50.2  & 64.1   \\ 
 \textbf{Chen et al.} \cite{DBLP:conf/ksem/ChenLWXWC20}   & \underline{35.7}  & \underline{26.5}  & \underline{54.1}   & 51.2  & 41.7  & 70.3  & 68.5  & 59.0  & \underline{86.9}  \\ 

 \textbf{HEA}~\cite{DBLP:journals/ijon/GuoTZZL21}& -  & 12.7  & 36.9   & -  & 26.2  & 58.1  & -  & 41.7  & 78.6 \\ \midrule
  \textbf{$\text{ACK-MMEA (ours)}$} & \textbf{38.7} ($\uparrow$3.0) & \textbf{30.4} ($\uparrow$3.9) & \textbf{54.9} ($\uparrow$0.8)   & \textbf{62.4} ($\uparrow$7.7) & \textbf{56.0} ($\uparrow$8.7) & \textbf{73.6} ($\uparrow$1.5)   & \textbf{75.2} ($\uparrow$3.1) &  \textbf{68.2} ($\uparrow$1.1) & \textbf{87.4} ($\uparrow$0.5)  \\
  
\midrule
\midrule

&  \multicolumn{3}{c}{\textbf{FB15K-YAGO15K (20\%)}} &  \multicolumn{3}{c}{\textbf{FB15K-YAGO15K (50\%)}}  &  \multicolumn{3}{c}{\textbf{FB15K-YAGO15K (80\%)}}  \\ 
\textbf{Methods}  & MRR(\%) & Hits@1 (\%) & Hits@10 (\%) & MRR (\%) & Hits@1 (\%) & Hits@10 (\%)   & MRR (\%) & Hits@1 (\%) & Hits@10 (\%)\\ 
\midrule
\textbf{TransE} \cite{DBLP:conf/nips/BordesUGWY13}   & 11.2   & 6.4   & 20.3   & 26.2 & 19.7 & 38.2  & 46.3  & 39.2 & 59.5 \\ 
\textbf{IPTransE}~\cite{DBLP:conf/ijcai/ZhuXLS17}    & 8.4   & 4.7   & 16.9  & 24.8 & 20.1 & 36.9    & 45.8  & 40.1 & 60.2 \\ 
\textbf{GCN-align} \cite{DBLP:conf/emnlp/WangLLZ18}   & 15.3  & 8.1  & 23.5    & 29.4  & 23.5  & 42.4     & 47.7  & 40.6  & 64.3  \\
 \textbf{SEA} \cite{DBLP:conf/www/PeiYHZ19}    & 21.8   & 14.1   & 37.1   & 38.8 &  29.4  &  57.7  &  60.5 & 51.4 & 77.3  \\  
 \textbf{IMUSE} \cite{DBLP:conf/dasfaa/HeLQ0LZ0ZC19}    & 14.2   &8.1   & 25.7   & 46.9 & 39.8 &  60.1  &  58.1 & 51.2 &  70.7\\
 \textbf{AttrGNN}~\cite{DBLP:conf/emnlp/LiuCPLC20} & 31.8  & 22.4  & 39.5   & 46.2  & 38.0  & 63.9  & 67.1 & \underline{59.9} &  78.7 \\ \midrule
 \textbf{PoE} \cite{DBLP:conf/esws/LiuLGNOR19}    & 15.4   & 11.3   & 22.9   & 41.4   & 34.7   & 53.6    & 63.5   & 57.3   & 74.6 \\ 
\textbf{PoE-rni} \cite{DBLP:conf/esws/LiuLGNOR19}   & \underline{33.4}   & \underline{25.0}   & \underline{49.5}    & \underline{49.8}   & \underline{41.1}   & \underline{66.9}   & 57.2   & 49.2   & 70.5  \\ 

   \textbf{Chen et al.} \cite{DBLP:conf/ksem/ChenLWXWC20} & 31.7  & 23.4  & 48.0   & 48.6  & 40.3  & 64.5  & \underline{68.2}  & 59.8  & \underline{83.9} \\
   \textbf{HEA}~\cite{DBLP:journals/ijon/GuoTZZL21} & -   & 10.5   & 31.3    & -   & 26.5   & 58.1  & -  & 43.3   & 80.1 \\\midrule
 
  \textbf{$\text{ACK-MMEA (ours)}$}   & \textbf{36.0} ($\uparrow$2.6)  & \textbf{28.9} ($\uparrow$3.9)  & \textbf{49.6} ($\uparrow$0.1)   & \textbf{59.3} ($\uparrow$9.5)  & \textbf{53.5} ($\uparrow$12.4)  & \textbf{69.9} ($\uparrow$3.0)  & \textbf{74.4} ($\uparrow$6.2) &\textbf{67.6} ($\uparrow$7.7) &  \textbf{86.4} ($\uparrow$2.5) \\
\bottomrule
\end{tabular}
}

\label{Main3}
\end{table*}

\section{Experiment}

\begin{table*}[t]
\centering
\caption{Variant experiments on FB15K-DB15K (80\%) and FB15K-YOGA15K (80\%).  “w/o” means removing corresponding module from the complete model. “repl.” means replacing corresponding module with the other module. The "$\downarrow$" means the value of performance degradation compared to the ACK-MMEA.
}
\resizebox{\linewidth}{!}{
\begin{tabular}{l|cccc}
\toprule
 & \multicolumn{4}{c}{\textbf{FB15K-DB15K (80\%)}}\\
\textbf{Variants} {  }{  }{  }{  } {  }{  }{  }{  }{  }{  }{  }{  } {  }{  }{  }{  } {  }{  }{  }{  } {  }{  }{  }{  } {  }{  }{  }{  } {  }{  }{  }{  } {  }{  }{  }{  } {  }{  }{  }{  } {  }{  }{  }{  } {  }{  }{  }{  } {  }{  }{  }{  } {  }{  }{  }{  } {  }{  }{  }{  } {  }{  }{  }{  } {  }{  }{  }{  } {  }{  }{  }{  } {  }{  }{  }{  } {  }{  }{  }{  }  &{  }{  }{  }{  }{  }{  }{  }{  } MRR (\%){  }{  }{  }{  } &{  }{  }{  }{  } Hits@1 (\%){  }{  }{  }{  }  & {  }{  }{  }{  } Hits@10 (\%){  }{  } &  $\triangle$ Avg (\%)  \\
\midrule
   \textbf{$\text{ACK-MMEA (ours)}$}  & \textbf{75.2} & \textbf{68.2} & \textbf{87.4}& - \\\midrule
{  }{  }{  }{  } \textbf{  w/o attribute uniformization}   & 73.2 ($\downarrow$2.0)  & 63.9 ($\downarrow$4.3)  & 82.8 ($\downarrow$4.6) &   $\downarrow$3.6   \\  
{  }{  }{  }{  } \textbf{  w/o attribute uniformization (Merge Operator)}  & 74.0 ($\downarrow$1.2)  & 65.1 ($\downarrow$3.1)  &83.5 ($\downarrow$3.9)   &   $\downarrow$2.7 \\  
{  }{  }{  }{  } \textbf{  w/o attribute uniformization (Generate Operator)}   & 74.2 ($\downarrow$1.0)  & 65.6 ($\downarrow$2.6)  & 84.1 ($\downarrow$3.3)  &   $\downarrow$2.3  \\
{  }{  }{  }{  } \textbf{  w/o text attribute }  & 73.4 ($\downarrow$1.8)  & 66.2 ($\downarrow$2.0)  &85.9 ($\downarrow$1.5)   &   $\downarrow$1.7 \\  
{  }{  }{  }{  } \textbf{  w/o image attribute }   & 72.5 ($\downarrow$2.7)  & 65.8 ($\downarrow$2.4)  & 86.2 ($\downarrow$1.2)  &   $\downarrow$2.1  \\\midrule
{  }{  }{  }{  } \textbf{  repl. GCN}    &  73.5 ($\downarrow$1.7) & 66.6 ($\downarrow$1.6) &  82.7 ($\downarrow$4.7) &   $\downarrow$2.6\\
{  }{  }{  }{  } \textbf{  repl. GAT}   &  74.1 ($\downarrow$1.1) & 65.5 ($\downarrow$2.7) & 83.1 ($\downarrow$4.3) &   $\downarrow$2.7 \\ 
{  }{  }{  }{  } \textbf{ w/o random dropouts}  & 72.7 ($\downarrow$2.5)  &  66.0 ($\downarrow$2.2) & 84.6 ($\downarrow$2.8) &   $\downarrow$2.5 \\ 
{  }{  }{  }{  } \textbf{  repl. random replacement}   & 71.2 ($\downarrow$4.0)  &  64.9 ($\downarrow$3.3) & 83.8 ($\downarrow$3.6) &   $\downarrow$3.6 \\\midrule
{  }{  }{  }{  } \textbf{  w/o attribute similarity loss}    & 74.0 ($\downarrow$1.2) & 68.1 ($\downarrow$0.1) & 85.7 ($\downarrow$1.7)  &   $\downarrow$1.0  \\ 
{  }{  }{  }{  } \textbf{  w/o neighbor dissimilarity loss}   & 73.6 ($\downarrow$1.6) & 67.7 ($\downarrow$0.5) & 86.9 ($\downarrow$0.5)  &   $\downarrow$0.8  \\
 
\midrule

\midrule
 & \multicolumn{4}{c}{\textbf{FB15K-YOGA15K (80\%)}}\\
\textbf{Variants}  &MRR (\%) & Hits@1 (\%)  & Hits@10 (\%) & $\triangle$ Avg (\%)  \\
\midrule
\textbf{$\text{ACK-MMEA (ours)}$}  & \textbf{74.4} & \textbf{67.6} & \textbf{86.4}& - \\\midrule
{  }{  }{  }{  } \textbf{  w/o attribute uniformization}   & 73.6 ($\downarrow$0.8)  & 64.2 ($\downarrow$3.4)  & 84.3 ($\downarrow$2.1) &   $\downarrow$2.1   \\  
{  }{  }{  }{  } \textbf{  w/o attribute uniformization (Merge Operator)}  & 73.5 ($\downarrow$0.9)  & 65.3 ($\downarrow$2.3)  &83.9 ($\downarrow$2.5)   &   $\downarrow$1.9 \\  
{  }{  }{  }{  } \textbf{  w/o attribute uniformization (Generate Operator)}   & 74.1 ($\downarrow$0.3)  & 66.0 ($\downarrow$1.6)  & 84.5 ($\downarrow$1.9)  &   $\downarrow$1.2  \\
{  }{  }{  }{  } \textbf{  w/o text attribute }  & 73.9 ($\downarrow$0.5)  & 65.8 ($\downarrow$1.8)  &84.7 ($\downarrow$1.7)   &   $\downarrow$1.3 \\  
{  }{  }{  }{  } \textbf{  w/o image attribute }   & 73.6 ($\downarrow$0.8)  & 65.7 ($\downarrow$1.9)  & 84.6 ($\downarrow$1.8)  &   $\downarrow$1.5  \\\midrule
{  }{  }{  }{  } \textbf{  repl. GCN}    &  73.0 ($\downarrow$1.4) & 66.3 ($\downarrow$1.3) &  84.3 ($\downarrow$2.1) &   $\downarrow$1.6\\
{  }{  }{  }{  } \textbf{  repl. GAT}   &  73.8 ($\downarrow$0.6) & 65.9 ($\downarrow$1.7) & 83.9 ($\downarrow$2.5) &   $\downarrow$1.6 \\ 
{  }{  }{  }{  } \textbf{ w/o random dropouts}  & 72.3 ($\downarrow$2.1)  &  65.7 ($\downarrow$1.9) & 84.1 ($\downarrow$2.3) &   $\downarrow$2.1 \\ 
{  }{  }{  }{  } \textbf{  repl. random replacement}   & 70.9 ($\downarrow$3.5)  &  64.2 ($\downarrow$3.4) & 83.1 ($\downarrow$3.3) &   $\downarrow$3.4 \\\midrule
{  }{  }{  }{  } \textbf{  w/o attribute similarity loss}    & 72.8 ($\downarrow$1.6) & 66.7 ($\downarrow$0.9) & 85.3 ($\downarrow$1.1)  &   $\downarrow$1.2  \\ 
{  }{  }{  }{  } \textbf{  w/o neighbor dissimilarity loss}   & 73.2 ($\downarrow$1.2) & 67.0 ($\downarrow$0.6) & 86.1 ($\downarrow$0.3)  &   $\downarrow$0.7  \\
 
\bottomrule
\end{tabular}
}

\label{Ablation}
\end{table*}

\subsection{Dataset}

We conducted experiments on FB15K-DB15K and FB15K-YAGO15K datasets~\cite{DBLP:conf/esws/LiuLGNOR19}, which are the two most popular datasets in MMEA. 
FB15K-DB15K is the entity alignment dataset\footnote{https://github.com/mniepert/mmkb} of FB15K and DB15K multi-modal knowledge graph, including 12,846 alignment seeds. FB15K-YAGO15K is the entity alignment dataset of FB15K and YAGO15K knowledge graphs, including 11,199 alignment seeds. As previous works~\cite{DBLP:conf/ksem/ChenLWXWC20, DBLP:journals/ijon/GuoTZZL21} recommended, we divided the two data sets into training and testing sets at 2:8, 5:5, and 8:2, respectively. 
We report the official MRR, Hits@1, and Hits@10 metrics for evaluation on different proportions of alignment seeds. 
For more details, please refer to the Appendix~\ref{sec:appe-dataset}.

\subsection{Comparision Methods}

We compare our method with six EA methods. They originally aggregate the text attribute and relation information. Here, we introduce the image attributes initialized by VGG16 for entity representation with the same aggregation manner of text attributes:
(1) \textbf{TransE}~\cite{DBLP:conf/nips/BordesUGWY13} assumes that the entity embedding $v$ should be close to the attribute embedding $a$ plus their relation $r$.
(2) \textbf{IPTransE}~\cite{DBLP:conf/ijcai/ZhuXLS17} is a translation-based method to jointly optimize entities and relations representation in knowledge graphs with an iterative and parameter sharing strategy. 
(3) \textbf{GCN-align} \cite{DBLP:conf/emnlp/WangLLZ18} transfers entities and attributes of per language to a common representation space through GCN. 
(4) \textbf{SEA}~\cite{DBLP:conf/www/PeiYHZ19} leverages labeled and unlabeled entities through adversarial training, and combines the image attributes.
(5) \textbf{IMUSE}~\cite{DBLP:conf/dasfaa/HeLQ0LZ0ZC19} uses a bivariate regression to merge the relations and multiple attributes.
(6) \textbf{AttrGNN}~\cite{DBLP:conf/emnlp/LiuCPLC20} divides KG into multiple subgraphs, effectively modeling various types of attributes.

Furthermore, we compare our method with four MMEA methods, which also do not introduce name attributes and focus on how to utilize the multi-modal attributes:
(7) \textbf{PoE}~\cite{DBLP:conf/esws/LiuLGNOR19} utilizes the image features and measures the credibility by matching the semantics of the entities to mining the relations.
(8) \textbf{PoE-rni}~\cite{DBLP:conf/esws/LiuLGNOR19} uses the relation, numeric literals and images attributes of PoE with the best performance. 
(9) \textbf{Chen et al.}~\cite{DBLP:conf/ksem/ChenLWXWC20} design a fusion module to integrate multi-modal attributes.
(10) \textbf{HEA}~\cite{DBLP:journals/ijon/GuoTZZL21} characterizes MMKG in hyperbolic space.

In contrast to these methods, our method generates a new attribute-consistent MMKG to uniform attribute information and learns more robust entity representations via ConsistGNN with a joint loss.


\subsection{Implementation Details}
For all baselines, we adopt the best hyper-parameters reported in their literature. For the EA baselines (1-6), we reproduce the performance through adding image attributes.
For the MMEA baselines (7-10), we copy the existing results reported in the literature~\cite{DBLP:conf/esws/LiuLGNOR19, DBLP:conf/ksem/ChenLWXWC20, DBLP:journals/ijon/GuoTZZL21}. 

Our model is implemented based on PyTorch, an open-source deep learning framework. 
The BERT version is bert-base-uncased in huggingface\footnote{https://github.com/huggingface/transformers} for text attributes initialization and VGG version is VGG16\footnote{https://github.com/machrisaa/tensorflow-vgg} for image attributes initialization.
The GNN (GCN and GAT) layer is 2, the training epoch is 200, the L2 regularization value is 0.0001, and the margin gramma value is 1.0. For hyper-parameters, the best random dropping rate $\rho$ is 0.35 and temperature coefficient $\tau$ is 0.5, and coefficients $\lambda_{1}, \lambda_{2}, \lambda_{3}$ are 5, 3 and 2. For the learning rate, we adopt the method of grid search with a step size of 0.001. The optimal learning rate is 0.001. 
All hyper-parameter settings are tuned on the validation data by the grid search with 5 trials. Refer to Appendix~\ref{sec:app-hyper} for more details. 
All experiments were conducted on a server with one GPU (Tesla V100).
The time analysis of our method is shown in Appendix~\ref{sec:app-runtime}.

\subsection{Main Results}

To verify the effectiveness of our ACK-MMEA, we report overall average results in
Table~\ref{Main3}. It shows performance comparisons on FB15K-DB15K and FB15K-YAGO15K datasets with different splits on training/testing data of alignment seeds, i.e., 2:8, 5:5, and 8:2. 

From the table, we can observe that: 1) Our attribute-consistent model outperforms all the baselines of both EA and MMEA methods, in terms of three metrics on both datasets. Specifically, our model improves $3.0\%-7.7\%$ ($4\%$ on average) on FB15K-DB15K and $2.6\%-9.5\%$ ($6\%$ on average) on FB15K-YAGO15K in terms of MRR for all proportions of training data, respectively. It demonstrates that our model is robust to different proportions of training resource, achieving reliable performance on multi-modal entity alignment.
2) Compared to EA baselines (1-4), especially for MRR and Hits@1, our model improves $5\%$ and $9\%$ up on average on FB15K-DB15K and FB15K-YAGO15K, tending to achieve more significant improvements. 
It demonstrates that effectiveness of multi-modal consistent-attribute uniformization for incorporating consistent alignment knowledge.
3) Compared to MMEA baselines (5-8), our model designs a ConsistGNN model on new attribute-consistent MMKGs, 
the average gains of our model regarding MRR, Hits@1 and Hits@10 are 5\%, 5\%, and 1\%, respectively. The reason is that our method incorporates the consistent multi-modal attributes and robust relation-aware entity information.
4) In terms of three proportions of training data on both datasets, our model improves $4.5\%$ on average and $8\%$ on average on FB15K-DB15K and FB15K-YAGO15K for the Hits@1 metric, which means the proportion that only prediction label is equal to the global label. It demonstrates that our method is more accurate compared to baselines, which can provide more correct predictions when only one outcome can be predicted.
All the observations demonstrate the effectiveness of the ACK-MMEA framework.

\subsection{Discussions for Model Variants}

\begin{figure}[t]
 \includegraphics[width=\linewidth]{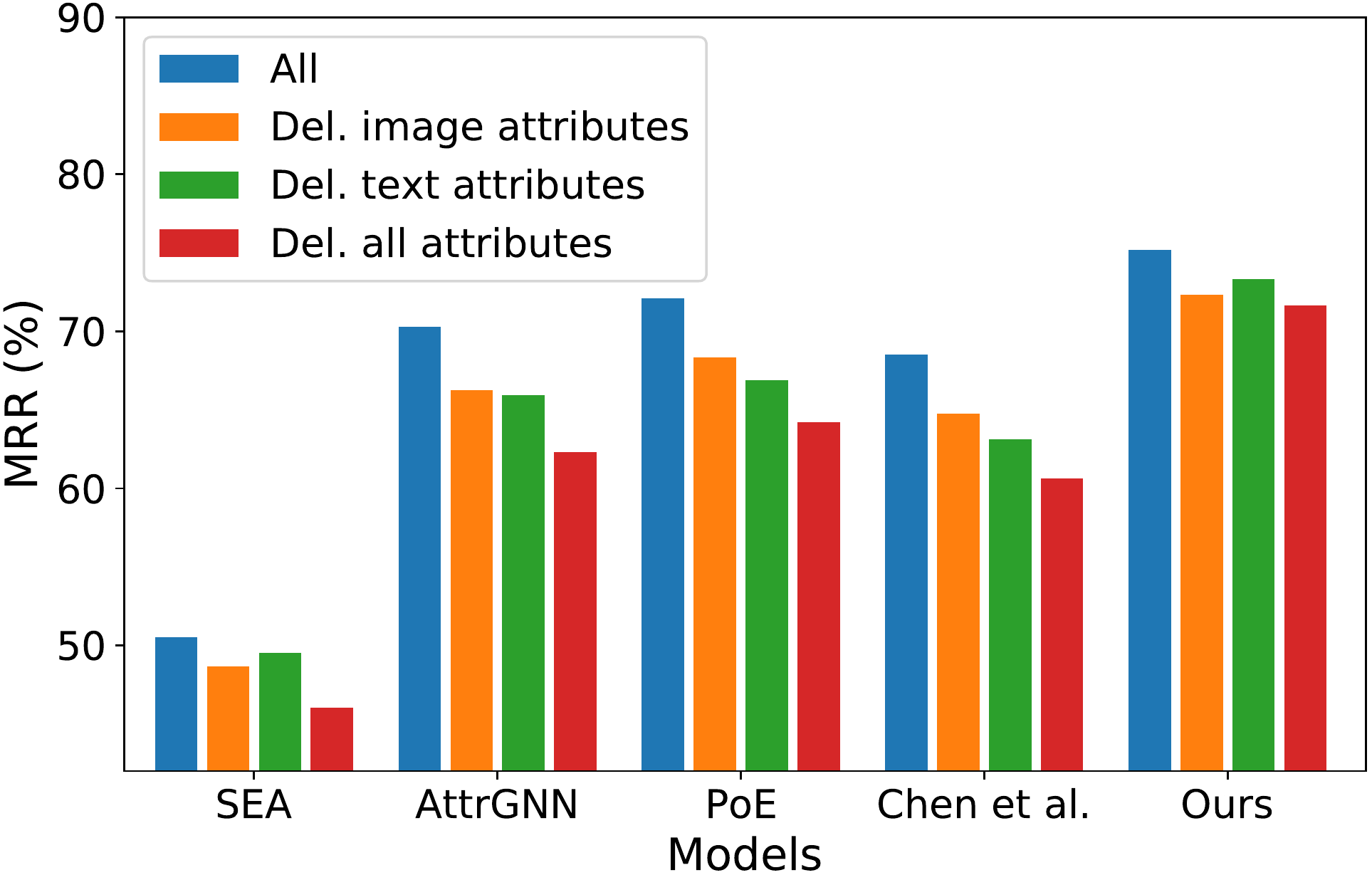}
    \caption{Results of deleting attributes on FB15K-DB15K (80\%). ``Del.'' means deleting the corresponding attribute.}
    \label{attributes}
\end{figure}

To investigate the effectiveness of each module in ACK-MMEA, we conduct variant experiments, showcasing the results in Table~\ref{Ablation} and Figure~\ref{attributes}. 
The "$\downarrow$" means the value of performance degradation compared to the ACK-MMEA.

From the Table~\ref{Ablation}, we can observe that: 1) The impact of the attribute uniformization tends to be more significant on using original attributes. We believe the reason is that the consistent attributes captures more clues for entity alignment. 2) By replacing the ConsistGNN to GCN, GAT or without random dropouts on neighbors, or random replacement on neighbors, the performance decreased significantly. It demonstrates that the ConsistGNN captures more effective consistent-attribute and relation information. 3) The impacts of the attribute similarity and neighbor dissimilarity loss tend to be significant. Since the consistent attributes tackle the contextual gap, and the neighbor loss guides our model to learn robust representations. 
4) When we remove all image attributes as ``w/o image attribute", our method drops 2.1\% and 1.5\% on average on FB15K-DB15K and FB15K-YAGO15K. The performance decreases 3.6\% and 2.1\% on average when we remove attribute uniformization module as ``w/o attribute uniformization''. It demonstrates that image attributes can improve model performance and our method utilizes image attributes effectively through capturing more alignment knowledge.
All the observations demonstrate the effectiveness of each component in our model.

To further investigate the impact of multi-modal attributes on all compared methods, we report the results by deleting different modality of attributes, as shown in Figure~\ref{attributes}. From the figure, we can observe that: 1) The variants without the text or image attributes significantly decline on all evaluation metrics, which demonstrates that the multi-modal attributes are necessary and effective for the entity alignment task. 2) Our model is less affected by deleting all multi-modal attributes. The reason we think is that the random dropouts on neighbors and the neighbor dissimilarity loss are beneficial to obtaining better entity representations. 3) Compared to other baseline methods, our model derives better results both in the case of using all multi-modal attributes or abandoning some of them. It demonstrates our model makes full use of existing multi-modal attributes, and consistent attributes are effective for the multi-modal entity alignment task. 4) When we delete all attributes as "Del. all attributes", our method drops 1\%-5\% in terms of MRR, and performs best compared to all baselines.
It demonstrates that our model makes the entity representation having fault tolerance through the relation-aware entity representation and more precise alignment of candidate entities by the dissimilarity constraint loss.
5) When we delete all image attributes as "Del. image attributes", which means that the original multi-modal knowledge graph transferred into a KG, our method is better than other baselines. It demonstrates that our model incorporates consistent alignment knowledge by the attribute-consistent relation representation encoder and the relation-aware entity representation encoder.
All the observations demonstrate that the effectiveness of the constructed attribute-consistent MMKG and the ConsistGNN.

\subsection{Impact of Dropping Rate}\label{Section 5.7}
\begin{figure}[t]
    \includegraphics[width=0.92\linewidth]{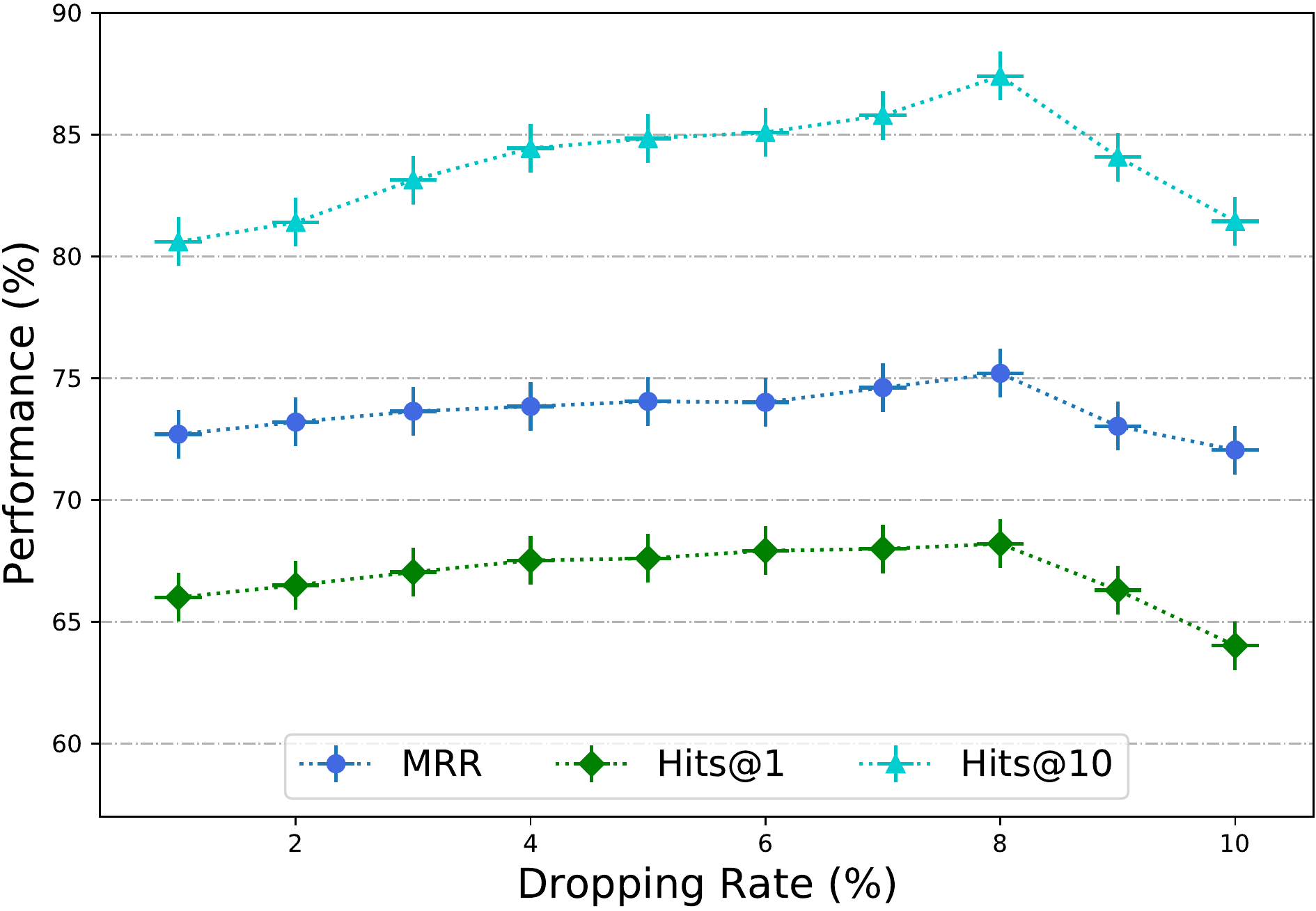}
    \centering
    \caption{Analysis of the dropping rate $\rho$.}
    \label{drop}
\end{figure}

We investigate the impact of the random dropouts rate on neighbors. 
Figure \ref{drop} shows the metric values with various hyper-parameter setting of $\rho$ on the FB15K-DB15K (80\%). As the dropping rate increases, the MRR, Hits@1 and Hits@10 gradually increase and then falling after an optimal value. The peak performance of the model is when the dropping rate of neighbors reaches 35\%, reflecting the effectiveness to learn the robust entity representation. It demonstrates the capacity of the random dropouts for improving the fault tolerance ability and enforcing consistent attribute aggregation.

\subsection{Impact of Attribute Number}

\begin{figure}[t]
    \includegraphics[width=0.92\linewidth]{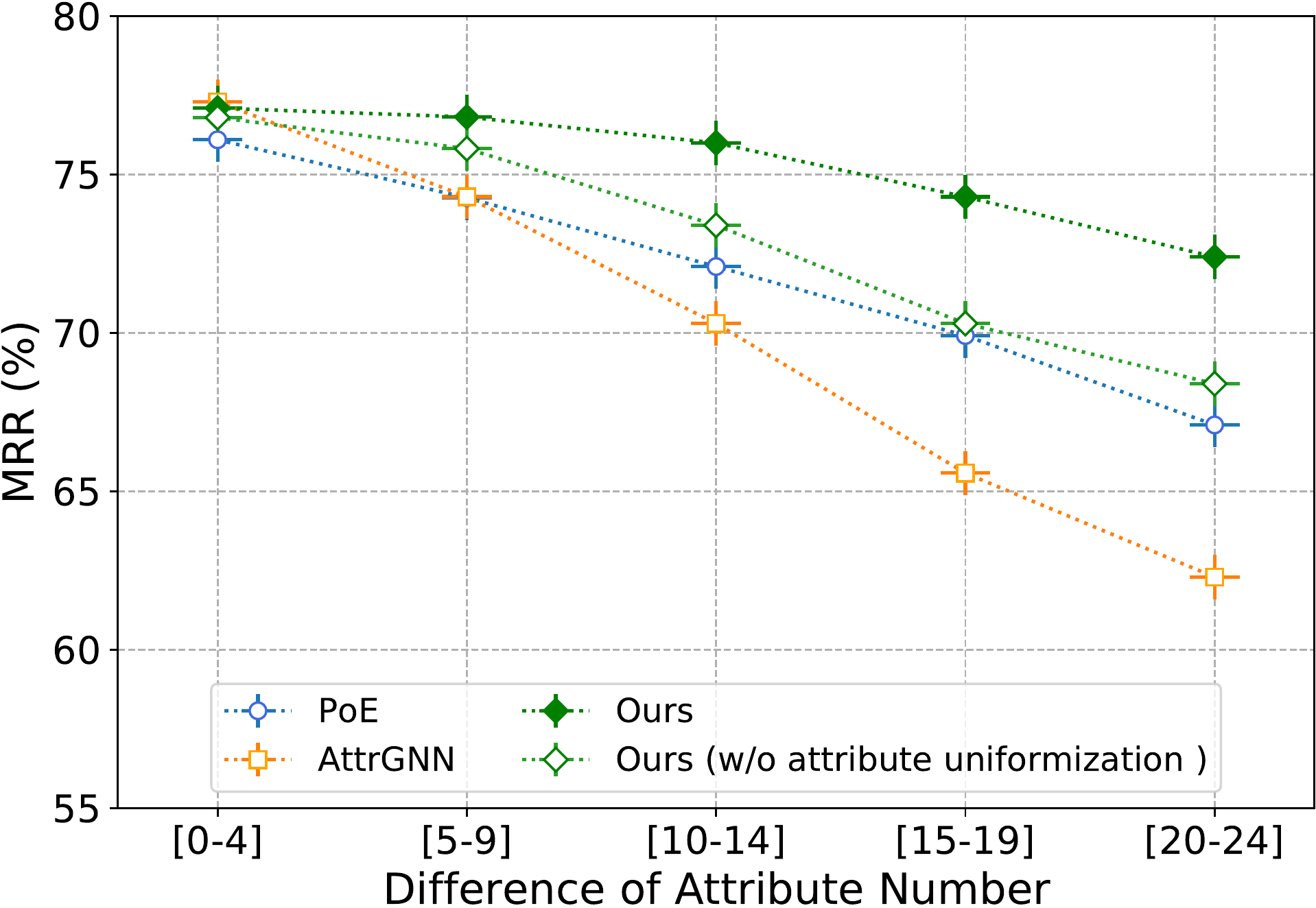}
    \centering
    \caption{Impact of differences in attribute number.}
    \label{difference}
\end{figure}

We investigate the impact of different degrees of the contextual gap between alignment seeds. 
To do so, we choose entity alignment seeds with the same number of image attributes, and vary the gaps of text attribute number of entity pairs in $[0,24]$.
Figure \ref{difference} shows the performance of different models in the case of varied attribute gaps on the FB15K-DB15K (80\%). From the figure, we can observe that: 1) With the increase of the gaps on alignment seeds, the performance of all methods gradually decreases. The main reason is that the bigger the gaps between entity seeds, the more difficult it is to match entities. This again confirms our intuition claimed in the introduction. 2) Compared to baseline methods, the performance of our model decreased slowly, demonstrating the superiority of our method in tackling the contextual gap issue. 
All the observations demonstrate that our method can reduce the impact of the contextual gap. 

\section{Conclusion}
This paper proposes a novel multi-modal entity alignment framework, namely ACK-MMEA. It generates an attribute-consistent MMKG with each entity containing only one attribute of each modality by the multi-modal attribute uniformization. We further propose the ConsistGNN to integrate the consistent multi-modal attributes and obtain aggregated relation representations and robust entity representations. 
To evaluate the attribute-consistent MMKG, we design the joint alignment loss with three objectives.
Our work overcomes the contextual gaps between entity pairs, caused by the information redundancy and absence of the attribute. 
The empirical experiments demonstrate that our method tackles the contextual gap problem. However, the operator of attribute generation will introduce noise data. 
In future work, we will study how to avoid the influence of noise data on the MMEA task.

\begin{acks}
We thank the anonymous reviewers for their insightful comments and suggestions. 
Jianxin Li is the corresponding author.
The authors of this paper were supported by the NSFC through grant No.U20B2053, 62106059 and the Academic Excellence Foundation of Beihang University for PhD Students. 

\end{acks}



\bibliographystyle{ACM-Reference-Format}
\bibliography{reference}


\clearpage

\appendix


\section{Related Work}
\label{sec:appe-Related}

\subsection{Entity Alignment}
Entity alignment (EA) technology mainly includes early embedding-based methods~\cite{DBLP:conf/nips/BordesUGWY13, DBLP:conf/ijcai/ChenTYZ17} and recently popular GNN-based methods~\cite{DBLP:conf/acl/CaoLLLLC19, DBLP:conf/emnlp/ShiX19}. The focus of various GNN-based methods is the aggregation way of attributes~\cite{DBLP:conf/acl/CaoLLLLC19, DBLP:conf/emnlp/LiuCPLC20}, relations~\cite{wang2018cross, DBLP:conf/ijcai/ChenTCSZ18}, and neighbor features~\cite{DBLP:conf/acl/WuLFWZ20, DBLP:conf/aaai/SunW0CDZQ20}. 
Specifically, the attribute-awared methods~\cite{DBLP:conf/ijcai/ChenTCSZ18, DBLP:conf/acl/CaoLLLLC19, DBLP:conf/ijcai/ZhangSHCGQ19} aggregate multi-type attributes or combine multiple models to encode entities for learning the entity embedding from multiple perspectives.
AttrGNN~\cite{DBLP:conf/emnlp/LiuCPLC20} divides KG into subgraphs for attributes aggregation, effectively modeling various types of attribute triples. 
The above methods demonstrate the effectiveness of aggregating attributes for EA.
Nevertheless, all of these methods ignore the inconsistency of attributes, as well as the image attributes.

\subsection{Multi-Modal Entity Alignment}
Furthermore, because of the multi-modal nature of KGs in real-world, there are several works~\cite{DBLP:journals/corr/abs-2202-05786,DBLP:conf/dsc/WangLG21, DBLP:conf/dsc/JiangLG21} beginning to focus on the MMEA technology.
Similar to EA on single-modal KG, many tasks on MMKG ~\cite{DBLP:conf/cikm/SunCZWZZWZ20, DBLP:conf/aaai/0001CRC21} provide the possibility of the fusion of multi-modal attributes and relations. 
As the first work on the MMEA, PoE~\cite{DBLP:conf/esws/LiuLGNOR19} characterized each entity as a single vector wherein all modality features of entities are concatenated. However, it cannot capture the potential interactions among heterogeneous modalities, limiting its capacity for performing accurate entity alignments. Later, \citet{DBLP:conf/ksem/ChenLWXWC20} proposed a multi-modal knowledge embedding method to discriminatively generate knowledge representations of different types of knowledge, and then designed a multi-modal fusion module to integrate them. \citet{DBLP:journals/ijon/GuoTZZL21} developed hyperbolic multi-modal entity alignment (HEA) approach to combine both attribute and entity representations in the hyperbolic space and used aggregated embeddings to predict alignments.
MCLEA~\cite{DBLP:journals/corr/abs-2209-00891} and MSNEA~\cite{DBLP:conf/kdd/ChenL00WYC22} reduce the gaps between modalities for each entity as well as utilize name embeddings.
Nevertheless, the above existing methods ignore contextual gaps between entity pairs and in turn may constrain the effectiveness of alignment.



\section{Datasets and Evaluation Metrics}
\label{sec:appe-dataset}

\subsection{Datasets}
In our experiments, we use two multi-modal datasets which are built in~\cite{DBLP:conf/esws/LiuLGNOR19}, namely FB15K-DB15K and FB15K-YAGO15K. FB15K is a representative subset extracted from the Freebase knowledge base. Aiming to maintain an approximate entity number of FB15K, DB15K from DBpedia and YAGO15K from YAGO are mainly selected based on the entities aligned with FB15K. 
Table~\ref{dataset} depicts the statistics of multi-modal datasets. 

\begin{table}[bp]
\caption{Statistics for the datasets. (Rel.: Relation, Attr.: Attribute, Rel. T.: Relational Triple, Attr. T.: Attributes triple.)}
\centering
\resizebox{\linewidth}{!}{
\begin{tabular}{l|rrrrrr}
\toprule
\textbf{Dataset} &  \#Entity &  \#Rel. & \#Attr. & \#Rel. T. & \#Attr. T. & \#Images \\ 
\midrule
 \textbf{FB15K} & 14,951 & 1,345  & 116 & 592,213  &29,395 &13,444 \\ 
 \textbf{DB15K} &12,842  & 279 & 225 &  89,197  &48,080  &12,837  \\ 
  \textbf{YAGO15K} & 15,404 & 32 & 7 & 122,886  & 23,532 & 11,194\\ 
\bottomrule
\end{tabular}
}
\label{dataset}
\end{table}

\subsection{Evaluation Metrics}
 We utilize cosine similarity to calculate the similarity between two entities and employ Hits@n, and MRR as metrics to evaluate all the models. Hits@n means the rate correct entities rank in the top n according to similarity computing. MRR denotes the mean reciprocal rank of correct entities. The higher values of Hits@n and MRR explain the better performance of the method.

\section{Hyper-parameters}\label{sec:app-hyper}

To enable replication and foster research, we report our hyper-parameter settings in Table~\ref{parameters}.
Note that all the hyper-parameter settings are tuned on the validation set by the grid search with 5 trials.
We adopt bert-large-uncased in huggingface as our encoder, whose layer number is 24 and the embedding size is 1024.
The best values of hyper-parameters $\lambda_{1}, \lambda_{2}, \lambda_{3}$ are 5, 3 and 2.
Specifically, if the model does not decrease the loss function of the validation set for 100 consecutive turns, the operation is stopped.
All baseline models use the same data set partitioning to ensure fairness.
To ensure fairness, all baselines use the same entity representation dimension, which is set to 128 dimensions. All experiments were conducted on a server with one GPU (Tesla V100).

\section{Empirical Runtime Analysis}\label{sec:app-runtime}
The time complexity of the proposed framework is acceptable. Table~\ref{Runtime} shows the time costs of the training of our method and three good performance baseline methods on FB15K-DB15K and FB15K-YAGO15K. For fairness, we only use one GPU, including AttrGNN. Thus, we train the three channels of AttrGNN in turn, which makes the model take the longest training time. 
Our method generates a attribute-consistent knowledge graph, which enhances the time cost. 
However, we design a random dropouts mechanism to save time. 
Overall, the time complexity of our approach can be on par with other efficient approaches.

\begin{table}[bp]\normalsize
\caption{Hyper-parameter settings of model.}
\centering
\resizebox{\linewidth}{!}{
\begin{tabular}{l|cc}
\toprule 
\textbf{Hyper-parameters}  & \textbf{FB15K-DB15K}  & \textbf{FB15K-YAGO15K} \\
\midrule
Batch size    &512  &512  \\
Train epoch   &200  &200   \\
Learning rate&0.001 &0.001  \\
Temperature&0.5&0.5\\
Negative Sample Number &15&15\\
Weight Decay &0.01&0.01\\
Random Dropouts Rate &0.35 &0.35  \\
$\lambda_{1}$ &5&5\\
$\lambda_{2}$ &3&3\\
$\lambda_{3}$ &2&2\\
\bottomrule
\end{tabular}}
\label{parameters}
\end{table}

\begin{table}[bp]
\caption{Average Training time (s) on the FB15K-DB15K and FB15K-YAGO15K datasets.}
\centering
\resizebox{\linewidth}{!}{
\begin{tabular}{l|cccc}
\toprule
\textbf{Methods} & AttrGNN & PoE &  PoE-rni &  ACK-MMEA \\ 
\midrule
 FB15K-DB15K & 396 & 164  & 162 & 165 \\ 
 FB15K-YAGO15K &389 & 155 & 151 &  156   \\ 
\bottomrule
\end{tabular}
}
\label{Runtime}
\end{table}

\end{document}